# НА ВХОДЕ ТЕКСТЫ, НА ВЫХОДЕ СМЫСЛ: НЕЙРОННЫЕ ЯЗЫКОВЫЕ МОДЕЛИ ДЛЯ ЗАДАЧ СЕМАНТИЧЕСКОЙ БЛИЗОСТИ (НА МАТЕРИАЛЕ РУССКОГО ЯЗЫКА)

**Кутузов А.** (akutuzov@hse.ru)
НИУ Высшая Школа Экономики
и Mail.ru Group, Москва, Россия

**Андреев И.** (i.andreev@corp.mail.ru)
Mail.ru Group, Москва, Россия

# TEXTS IN, MEANING OUT: NEURAL LANGUAGE MODELS IN SEMANTIC SIMILARITY TASKS FOR RUSSIAN

**Kutuzov A.** (akutuzov@hse.ru)
National Research University Higher School of Economics and Mail.ru Group, Moscow, Russia

**Andreev I.** (i.andreev@corp.mail.ru)
Mail.ru Group, Moscow, Russia

Distributed vector representations for natural language vocabulary get a lot of attention in contemporary computational linguistics. This paper summarizes the experience of applying neural network language models to the task of calculating semantic similarity for Russian. The experiments were performed in the course of Russian Semantic Similarity Evaluation track, where our models took from 2nd to 5th position, depending on the task.

We introduce the tools and corpora used, comment on the nature of the evaluation track and describe the achieved results. It was found out that Continuous Skip-gram and Continuous Bag-of-words models, previously successfully applied to English material, can be used for semantic modeling of Russian as well. Moreover, we show that texts in Russian National Corpus (RNC) provide an excellent training material for such models, outperforming other, much larger corpora. It is especially true for semantic relatedness tasks (although stacking models trained on larger corpora on top of RNC models improves performance even more).

High-quality semantic vectors learned in such a way can be used in a variety of linguistic tasks and promise an exciting field for further study.

## 1. Introduction

This paper describes authors' experience with participating in Russian Semantic Similarity Evaluation (RUSSE) track. Our system was trained using neural network language models; the process is explained below, together with the workflow for evaluation. We also comment on the nature of the RUSSE tasks and discuss features of neural models for Russian.

Since Ferdinand de Saussure, it is known that linguistic sign (including word) is arbitrary. It means that there is no direct connection between its form and concept (meaning). Consequently, printed orthographic words *per se* do not contain sense. What is important for the task discussed here, is that if given only disjoint word forms, a computer (an artificial intelligence) can't hope to grasp the concepts behind them and decide whether they are semantically similar or not.

At the same time, detecting degree of semantic similarity between lexical units is an important task in computational linguistics. The reason is threefold. First, it is a means in itself: often, applications demand calculating the "semantic distance" between words, for example, in finding synonyms or near-synonyms for search query expansion or other needs [Turney and Pantel 2010]. Second, once we know which words are similar and to what extent, we can "draw a semantic map" of the language in question and use this knowledge in a multitude of tasks, from machine translation [Mikolov et al. 2013b] to natural language generation [Dinu and Baroni 2014]. Finally, measuring performance in semantic similarity task is a convenient way to estimate soundness of a semantic model in general.

Consequently, various methods of overcoming linguistic arbitrariness and calculating semantic similarity for natural language texts were invented and evaluated for many widespread languages. However, computational linguistics community lacks experience in computing semantic similarity for Russian texts. Thus, the task of applying state-of-the-art methods to this material promised to be interesting, and kept its promise.

The paper is structured as follows. In the Section 2 we give a brief outline of RUSSE evaluation track. The Section 3 describes the models we used to compute semantic similarity and the corpora to train these models on. In the Section 4, results are evaluated and influence of various model settings discussed. The Section 5 lists the main results of our research. In the Section 6, we conclude and propose directions for future work.

## 2. Task Description

RUSSE[1] is the first attempt at semantic similarity evaluation contest for Russian language. It consists of four tracks: two for the relatedness task and two for the association task. Participants were presented with a list of word pairs and had to fill in the degree of semantic similarity between each pair, in the range [0;1].

---

[1] http://russe.nlpub.ru; the authors of the present paper are under the number 9 in the participants' list.

In the semantic relatedness task, participants were to detect word pairs in synonymic, hyponymic or hypernymic relations and to separate them from unrelated pairs. First track test set in this task included word pairs with human-annotated similarities between them. Systems' performance was measured with Spearman's rank correlation between these human scores and the system scores. The second track aim was to distinguish between semantically related pairs from RuThes Lite thesaurus [Лукашевич 2011] and random pairings. Average precision was used as evaluation metrics for this track and for the tracks in the second task.

In the association task, participants had to detect whether the words or multi-word expressions are associated (topically related) to each other. First track in this task mixed random pairings and associations taken from the Russian Associative Thesaurus[2]. The second track test set included associations from Sociation.org database[3].

An ideal system should have always assigned 0 to unrelated pairs and positive values to related or associated ones, thus achieving average precision of 1.0. In the case of the first semantic relatedness track an ideal system was to rank the pairs identically to the human judgment, to achieve Spearman's rho of 1.0.

In the end, participants were rated with four scores: **hj** (Spearman's rho for the first relatedness track), **rt** (average precision for the second relatedness track), **ae** (average precision for the first association track) and **ae2** (average precision for the second association track). The contest itself is described in detail in [Panchenko et al. 2015]. We participated in all tracks, using different models.

In general, the choice of test data and evaluation metrics seems to be sound. However, we would like to comment on two issues.

1. Test sets for the **rt** and **ae2** tasks include many related word pairs which share long character strings (e.g., "*благоразумие; благоразумность*"). This allows reaching unexpectedly good performance without building any complicated models, using only character-level analysis. We were able to achieve average precision of 0.79 for **rt** task and 0.72 for **ae2** task with the following algorithm: if two words share strings more than 3 characters in length, choose the longest of such strings; its length divided by 10 is the semantic similarity between words; if no such strings are found, assume similarity is zero.

    It seems trivial that in Russian, words which share stems are virtually always semantically similar in this or that way. Thus, the contest would benefit if the ratio of such pairs became lower, so that the participants had to design systems that strive to understand meaning, not to compare strings of characters. Certainly, this issue is conditioned by the usage of RuThes and Sociation databases, which by design contain lots of related words with common stems. It is difficult to design a dataset of semantically related lexical units for Russian which would not be haunted by this problem. However, this is the challenge for organizers of the future evaluations. Other RUSSE tracks do not suffer from this flaw.

---

[2] http://tesaurus.ru/dict/dict.php

[3] http://sociation.org

2. The test set for the **ae** task was Russian Associative Thesaurus. It was collected between 1988 and 1997; many entries can already be considered a bit archaic (*"колхоз; путь ильича"*, *"президент; ельцин"*, etc). Perhaps, this is the reason for often observed disagreement in systems' performance measured in **ae** and in **ae2**. These datasets differ chronologically, and it greatly influences association sets. Note striking difference in comparison to semantic relatedness task: synonymic, hyponymic and hypernymic relations are stable for dozens or even hundreds of years, while associations can dramatically change in ten years, depending on social processes. At the same time, such glitches cover only small part of the entries, and this is only a minor remark.

In the next chapter we describe our approach to computing semantic similarity for Russian.

## 3. Neural Networks Meet Corpora

The methods of automatically measuring semantic similarity fall into two large groups: knowledge-based and distributional ones [Harispe et al. 2013]. The former depend on building (manually or semi-automatically) a comprehensive ontology for a given language, which functions as a conceptual network. Once such a network is complete, one can employ various measures to calculate distance between concepts in this network: in general, the shorter is the path, the higher is the similarity.

We employed other, distributional approach, motivated by the notion that meaning is defined by usage and semantics can be derived from the contexts a given word takes [Lenci 2008]. Thus, these algorithms are inherently statistical and data-driven, not ruled by a curated conceptual system, as is the case for knowledge-based ones.

If lexical meaning is generally the sum of word usages, then the most obvious way to capture it is to take into account all contexts a word participates in, given a large enough corpus. In distributional semantics, words are usually represented as vectors in semantic space [Turney and Pantel 2010]. In other words, each lexical unit is a vector of its "neighborhood" to all other words in the lexicon, after applying various distances and weighting coefficients. The matrix of *n* rows and *n* columns (where *n* is the size of the lexicon) with "neighborhood degrees" in the cells is then a distributional model of the language. One can compare vectors for different words (e.g., calculating their cosine similarity) and find how "far" they are from each other. This distance turns out to be the semantic similarity we sought, expressed continuously from **0** (totally unrelated words) to **1** (absolute synonyms).

Such an approach theoretically scales well (one has to simply add more texts to the corpus to get new words and contexts) and does not demand laborious and subjective process of building an ontology. Meaning is extracted directly from linguistic evidence: the researcher only has to polish weighting algorithms. Also, fixed-length vector representations instead of orthographic words constitute excellent input to machine learning systems, independent of their particular aim.

The fly in the ointment is that traditional distributional semantic models (DSMs) are very computationally expensive. The reason is the dimensionality of their vectors, generally equal to the size of the lexicon. As a result, a model has to operate on sparse but very large matrices. For example, if a corpus includes one million distinct word types (not a maximum value, as we show below), we will have to compute dot products of 1M-dimensional vectors each time we need to find how similar two words are. Vectors' dimensionality can be reduced to reasonable values using tricks like singular value decomposition or principal components, but this often degrades performance or quality.

As a kind of remedy to this, artificial neural networks can learn distributed vector representations or "neural embeddings" of comparatively small size (usually hundreds of components) [Bengio 2003]. Neural models are directly trained on large corpora to produce vectors which maximize similarity between contextual neighbors found in the data, while minimizing similarity for unseen contexts. Vectors are initialized randomly, but in the course of the training the model converges and semantically similar words obtain similar vector representations. However, these models were slow to train because of non-linear hidden layer.

Recently, **Continuous Bag-of-Words** (CBOW) and **Continuous Skip-gram** neural network language models without hidden layer, implemented in the *Word2Vec* tool [Mikolov et al. 2013a], seriously changed the field; using smart combination of already known techniques, they learn high quality embeddings in a very short time. These algorithms clearly outperform traditional DSMs in various semantic tasks [Baroni et al. 2014].

For this competition, we tested both CBOW and skip-gram models. Evaluation results (for a wide range of settings) are given in Section 4.

In order to train neural language models one needs not only algorithms, but also corpora. We used 3 text collections:

1. **News**: a corpus of contemporary Russian news-wire texts collected by a commercial news aggregator. Corpus volume is about 1.8 billion tokens, more than 19 million word types. It was crawled from 1500 news portals, and news pieces themselves are dated from 1 September of 2013 to 30 June of 2014 (more than 9 million documents total).
2. **Web**: a corpus of texts found on Russian web pages. It originates from a search index for one of the major search engines in the Russian market, thus is supposed to be quite representative. This source repository itself contains billions of documents, but to train the model we randomly selected about 9 million pieces (no attention was paid to their source or any other properties). Thus, hopefully the corpus contains all major types of texts found in the Internet, in nearly all possible genres and styles.
Boilerplate and templates were filtered out to leave only main textual content of these pages, with the help of *boilerpipe* library [Kohlschütter et al. 2010]. After removing non-Cyrillic sentences, the resulting web corpus contained approximately 940 million tokens.
3. **Ruscorpora**: Russian National Corpus consists of texts which supposedly represent the Russian language as a whole. It has been developed for more than 10 years by a large group of top-ranking linguists, who select texts and segments for inclusion into the corpus. It was extensively described in the literature (see [Плунгян 2005], [Савчук 2005]). The size of the main part of RNC is 230 million word tokens, but we worked with the dump containing 174 million tokens.

All the corpora were lemmatized with *MyStem* [Segalovich 2003]. We used version 3.0 of the software, with disambiguation turned on. Stop-words were removed, as well as single-word sentences (they are useless for constructing context vectors). Because we removed stop-words ourselves, *word2vec* sub-sampling feature was not used. After this pre-processing, **News** corpus contained 1,300 million tokens, **Web** corpus 620 million tokens, and **Ruscorpora** 107 million tokens.

These corpora represent three different "stimuli" to neural network training algorithm. **Ruscorpora** is a balanced academic corpus of decent but comparatively small size, **Web** is large, but noisy and unbalanced. Finally, **News** is even larger than **Web**, but cleaner and biased towards one particular genre. These differences caused different results in semantic similarity tasks for models trained on the corpora in question (although all corpora proved to be good training sets).

We note that **Ruscorpora**, notwithstanding its size, certainly won this race, receiving scores essentially higher than the models trained on other two collections. The details are given in the next section.

## 4. Evaluation

There can be two reasons for a model to perform worse in comparison to the gold standard in this evaluation contest: either the model outputs incorrect similarity values (cosine distances in our case), or one or both words in the presented pair are unknown to the model. The former can be treated only by re-training the model with different settings or different training set, while the latter can be partially remedied by a couple of tricks, both of which we used.

The first trick exploits the issue described in the Section 2: many semantically similar words in Russian have common stems. We "computed" similarity using the longest common string algorithm in case of unknown words, as a kind of "emergency treatment". For **Ruscorpora** models it consistently increased average precision in **rt** track by 0.02…0.05.

Another trick is building model assemblies, allowing to "fall back" to another model in case when unknown words are met. In our case, we knew that **Ruscorpora** model is the best, but only for the words it knows. The **Web** model is slightly worse, but knows a lot more distinct words (millions instead of hundreds of thousands). Thus, we query **Web** model for the word pairs unknown to **Ruscorpora**. Similarity measures range strictly from 0 to 1 and are generally compatible across models. Only if the words are unknown even to the **Web** model, we fall back further to the longest common string trick. In our experience, such assemblies seriously improved overall performance.

Most important training parameters for our task are algorithm, vector size, window size and frequency threshold. The algorithm can be either CBOW or skip-gram, with the latter being considerably slower. Also, skip-gram performance was consistently worse for all corpora except **news**. This seems to be specific for Russian, as previous research for English corpora stated that skip-gram is generally better [Mikolov 2013a].

Vector size is the number of dimensions in vector representations; increasing vector size generally increases both performance and training time. Window is context width: how many words to the right and to the left will be considered. Larger window size increases training time and also leads to the model being more "topical" opposed to "functional" [Levy and Goldberg 2014]. It means that the model assigns similar vectors to topically associated words, not only to direct semantic relatives (synonyms, etc). This is quite natural, as the model trains on neighbors more distant from the analyzed lexical units. Unsurprisingly, models trained on large windows perform better in association tasks, while those trained on micro-windows of size 1 or 2 (only immediate neighbors) excel at catching direct semantic or functional relations.

Finally, frequency threshold or minimal count is a minimum frequency a word must possess in order to be considered by the model. All the lexical units with lower frequency are ignored during training and are not assigned vector representations. It is useful in order to get rid of low-frequency noise and train only on sufficiently presented evidence. Moreover, the less distinct words the model possess, the faster is training; the downside is, of course, absence of some words in the model lexicon.

In our experience, typical training speed on an Intel Xeon E5620 2.4GHz machine (14 cores) was 116,386 words per second for CBOW algorithm. Web corpus model training with vector size 500, minimal count 100, window 10 and 5 iterations (epochs) took approximately 7 hours; the model saw 3 168 819 885 words in total. This timing is consistent with [Mikolov et al. 2013a].

The Table 1 presents our best-performing models, as submitted to RUSSE contest.

**Table 1.** Our best results submitted to the evaluation

| Track | hj | rt | ae | ae2 |
| --- | --- | --- | --- | --- |
| Rank (among 18 participants) | 2 | 5 | 5 | 4 |
| Training settings | CBOW on **Ruscorpora** with context window 5, minimal count 5 + CBOW on **Web** with context window 10, minimal count 2 | CBOW on **Ruscorpora** with context window 5, minimal count 5 + CBOW on **Web** with context window 10, minimal count 2 | Skip-gram on **News** with context window 10, minimal count 10 | CBOW on **Web** with context window 5, minimal count 2 |
| Score | 0.7187 | 0.8839 | 0.8995 | 0.9662 |

Note that minimal count values (defining how much of low-frequency long tail is cut off) are different for different corpora. The optimal setting possibly depends on the vocabulary distribution in a particular text collection, and on how closely it follows Zipfian law. We leave this for further research.

It is clear that **Ruscorpora** beats both **Web** and **News** corpus in the task of distinguishing semantically related words. This is impressive considering its size: it seems that balance and clever selection of texts for corpus do really make sense and allow the model to learn very high quality vectors. However, when we turn to the task of detecting associations, sheer volume and diversity of **News** and **Web** become paramount, and they outperform **Ruscorpora** models. It is interesting that **News** model is better with predicting associations from Russian Associative Thesaurus. Probably, this reflects more "official" spirit of this resource in comparison with more colloquial nature of Sociaton.org database in the **ae2** track, better modeled with **Web** texts.

The plots below show how performance in different RUSSE tracks depends on training settings. Two parameters did not change: training mode (CBOW for **Ruscorpora** and **Web** and skip-gram for **News**) and minimal count (5 for **Ruscorpora**, 2 for **Web** and 10 for **News**); they reproduce the values in the Table 1. Only selected plots are shown here; see the link to the others in the Section 5.

The plots prove that while increasing vector size generally leads to quality increase, after a certain threshold this growth can sometimes stop or even revert[4]. This is the case for **Ruscorpora** (*Fig. 1*), but not for **Web** (*Fig. 2*) or **News**. We hypothesize that the reason is the size of these two corpora: the volume of data allows filling vector components with meaningful relationships, while with **Ruscorpora** the model can't learn so many relationships because of data insufficiency; as a result, vectors are filled with noise. This is again consistent with the notion that vector size increase must be accompanied by data growth, expressed in [Mikolov et al. 2013].

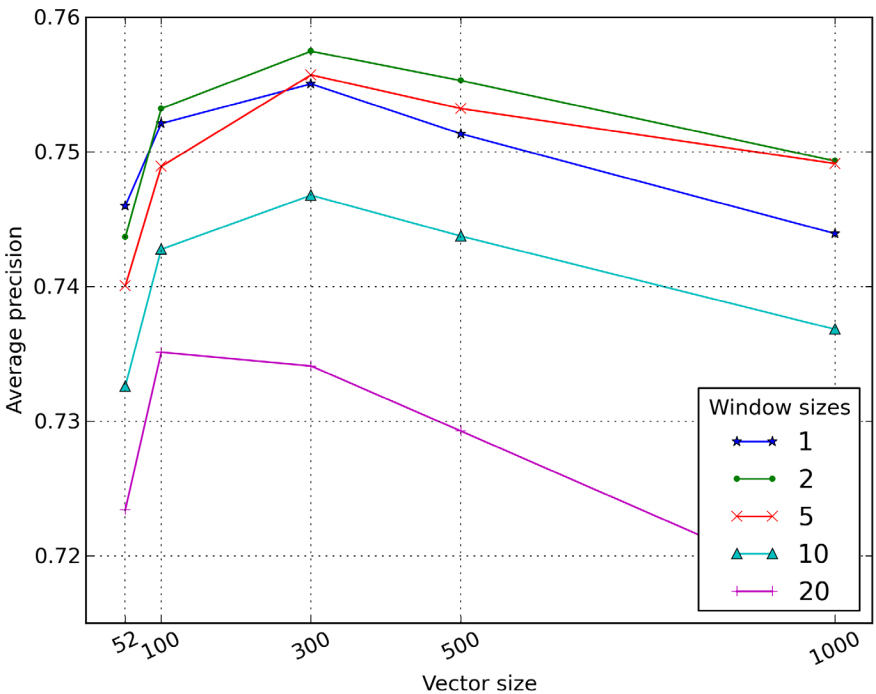

**Fig. 1.** Ruscorpora model performance in **rt** track depending on vector size

---

[4] Vector sizes start with 52, because training time is optimal when dimensionality is a multiple of 4.

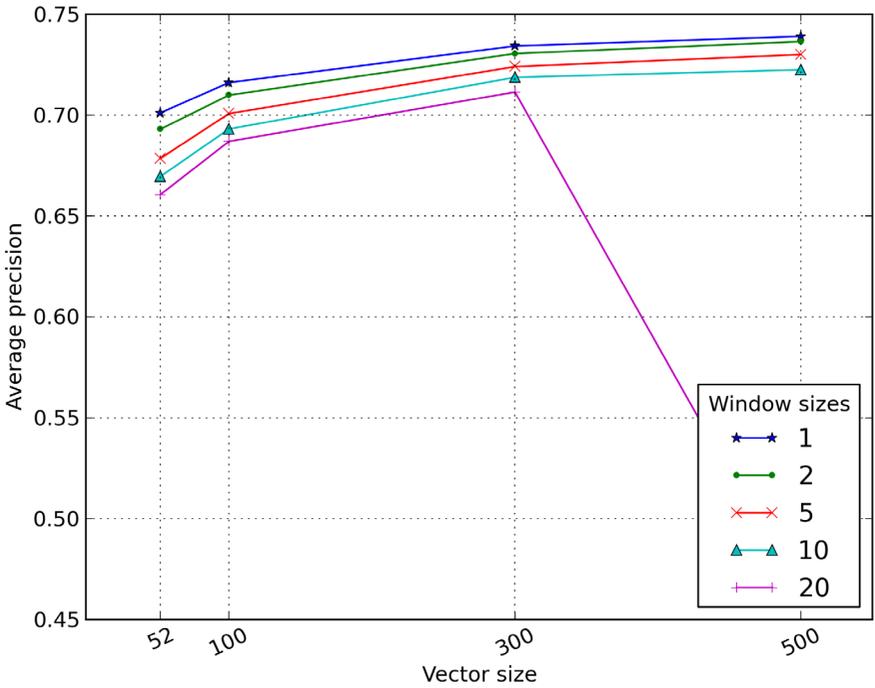

**Fig. 2.** Web model performance in **rt** track depending on vector size

As for the window size dynamics, we observe clear direct correlation between window size and **ae2** performance and inverse correlation for **rt** performance (*Fig. 3*). As already stated, a shorter window favors strict functional and semantic relations, while a larger window (10 words and more) allows catching more vague topical relations. Interestingly, **Ruscorpora** models are better at **ae** task with short windows, unlike **ae2** (*Fig. 4*); perhaps, associations from **ae** dictionary are more syntagmatic and tend to occur close to each other, while Sociation pairs are topical *par excellence*. This further proves deep difference between these two associative tasks.

## 5. Discussion

The first result of our research is that neural embedding models are shown to be directly applicable to Russian semantic similarity tasks. Rich morphology does not pose an obstacle for learning meaningful vector representations, with preprocessing limited to lemmatizing (training on unlemmatized text decreases performance, unlike English tasks where one often doesn't need to even stem the corpus). The result is very persuasive. We believe it is worth to try augmenting many NLP tools for Russian with neural embeddings to make existing instruments more semantically aware.

Kutuzov A., Andreev I.

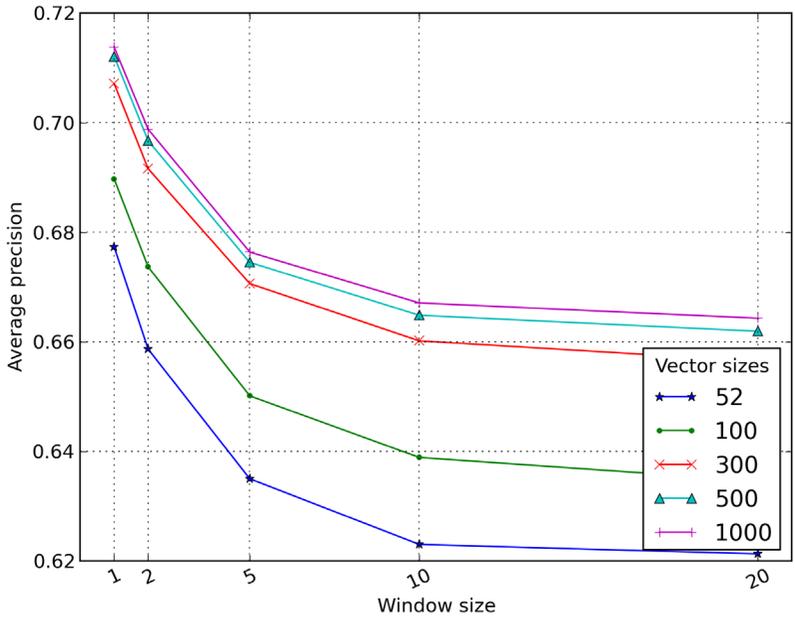

**Fig. 3.** News model performance in **rt** track depending on window size

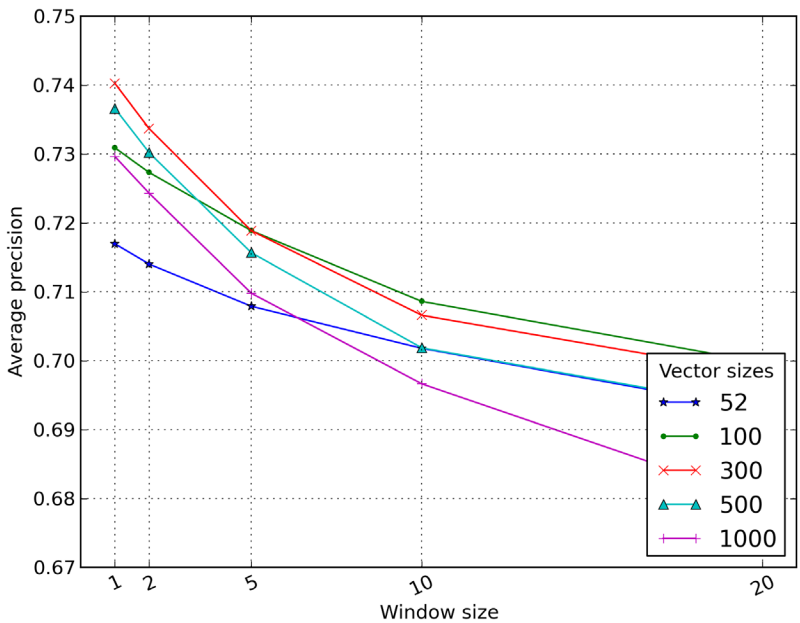

**Fig. 4.** Ruscorpora model performance in **ae** track depending on window size

Another, more unexpected outcome of our participation in RUSSE was that Russian National Corpus (RNC) turned out to be an excellent training set for neural network language models. When at start, we were sure that the amount of data plays dominant role and that the national corpus will eventually lose, because of being substantially smaller. However, it was quite the opposite: in the majority of comparisons (especially for semantic relatedness task) models trained on RNC outperformed their competitors, often even with vectors of lower dimensionality.

The only explanation is that RNC is really representative of the Russian language, thus providing balanced linguistic evidence for all major vocabulary tiers. Additionally, it seems to contain little or no noise and junk fragments, which sometimes occur in other corpora. To sum it up, we certainly recommend training neural language models on RNC, if this resource is available.

The resulting models for each of the three corpora, trained with optimal settings, can be downloaded at http://ling.go.mail.ru/misc/dialogue_2015.html; the full set of performance plots for different training settings is also there.

## 6. Future Work

We have only scratched the surface of exploiting neural embeddings to deal with Russian language material. The next step should be to perform a comprehensive study of errors typical for each model in their semantic similarity or other decisions. This can shed light on the real nature of differences between models and help in studying human errors.

Another very interesting field of research is corpora comparison through the output of neural language models trained on them [Kutuzov and Kuzmenko 2015]. Here we, in a way, arrive to an almost omnipotent "mind" able to rapidly evaluate huge corpora, taking into consideration what meanings words in their vocabularies take and how they are different from each other.

Of course, this is not an exhaustive outlook of computational linguistics research directions related to neural lexical vectors. Their foundational nature allows to employ them everywhere meaning is important; we anticipate a serious growth in semantic tools' quality.

Last but not least, we plan to implement a full-fledged web service for testing and querying distributed semantic models for Russian, particularly neural ones. A prototype to try with is already available online at http://ling.go.mail.ru/dsm.

### Acknowledgments

The authors thank the anonymous reviewers for their helpful comments. Support from the Basic Research Program of the National Research University Higher School of Economics is also acknowledged.